\documentclass[sigconf]{acmart-me}

\usepackage{booktabs} 
\usepackage{url}
\usepackage{color}
\usepackage{enumitem}
\usepackage{array}
\usepackage{algorithm,algorithmic}
\setcopyright{rightsretained}

\acmDOI{}

\acmISBN{}

\acmConference[MediaEval'21]{Multimedia Evaluation Workshop}{December 13-15 2021}{Online} 
\acmYear{2021}
\copyrightyear{}

\acmPrice{}

\begin{document}
\title{Two Stream Network for Stroke Detection in Table Tennis}

\author{Anam Zahra, Pierre-Etienne Martin}
\affiliation{CCP Department, Max Planck Institute for Evolutionary Anthropology, D-04103 Leipzig, Germany}
\email{anam_zahra@eva.mpg.de, pierre_etienne_martin@eva.mpg.de}

%
%
%
%
%

\renewcommand{\shorttitle}{Sports Video Task}

\begin{abstract}
This paper presents a table tennis stroke detection method from videos. The method relies on a two-stream Convolutional Neural Network processing in parallel the RGB Stream and its computed optical flow. The method has been developed as part of the MediaEval 2021 benchmark for the Sport task. Our contribution did not outperform the provided baseline on the test set but has performed the best among the other participants with regard to the mAP metric.
\end{abstract}

\begin{teaserfigure}
    \includegraphics[width=\linewidth]{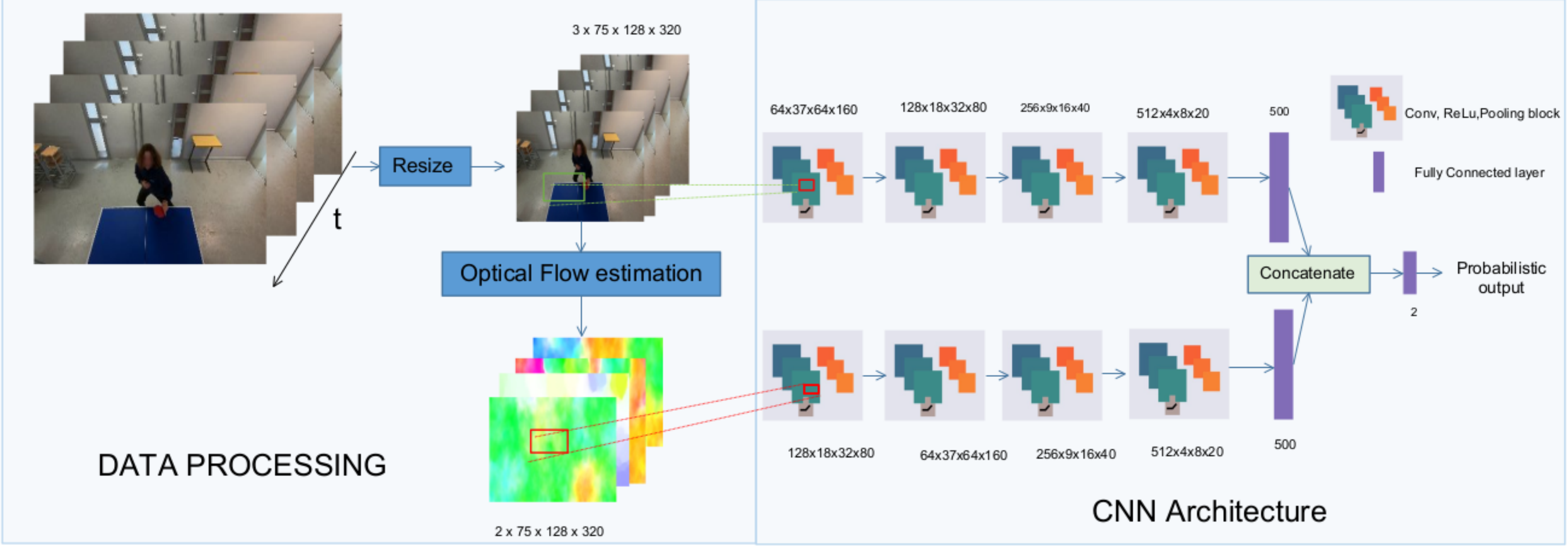}
    \caption{Pipeline method for stroke detection from videos. Cuboids of RGB and optical flow are fed to the network and classified as stroke or non-stroke. The feature dimension is described as follow: $RGB channels \times temporal \times height \times width$.
    }
    \label{fig:model_fig}
\end{teaserfigure}
%
%
%
%
%


\maketitle

\section{Introduction}
\label{sec:intro}

With the advent of Convolutional Neural Networks (CNNs), especially after the success of AlexNet~\cite{krizhevsky2012imagenet}, object detection, localization, and classification from images and videos have greatly progressed~\cite{krizhevsky2012imagenet,girshick2015fast,wang2013action,carreira2017quo}.The development of computer vision methods has motivated broader applications in the academic world. Our team is currently working on egocentric recordings from children in kindergarten and at home. The analysis of these recordings shall give us an automatic overview of their interactions on a daily basis. We hope to link these interactions with their cognitive development and, thereby, better understand early child development. With our participation in the Sports Video Task~\cite{mediaeval/Martin21/task}, in the stroke detection subtask, we hope to perfect our knowledge in event detection and transpose it to our project.
\par
The diversity of applications and visual data in sport, makes sports video analysis attractive for researchers. Automated sport event detection and action classification, especially from low-resolution videos, are helpful for monitoring and training purposes. For example in~\cite{hughes2002use,lees2003science}, the authors automate the performance analysis for the training optimization of players. Similarly, Sports Video task at MediaEval 2021 benchmark aims at improving athlete performance and training experience through the first steps of stroke detection and classification from videos.
\par
Event detection in videos is the first step to many other hot-topics such as video summarizing~\cite{khan2015video}, automated semantic segmentation~\cite{ballan2011event} and action recognition~\cite{dhamsania2016survey,martin2020fine}. These methods may be used to build summary, selecting highlights, and assisting players in training sessions. One way to approach the problem of event detection in sports with balls, can be through ball detection and tracking. Several researchers have tried to get the 2D, and 3D ball trajectories in order to achieve so~\cite{tamaki2013reconstruction,myint2015tracking,myint2016tracking}.
\par
Inspired from~\cite{koch2015siamese,PeICIP19,PeICPR20,voeikov2020ttnet}, this method combines the optical flow and features learned from the RGB stream in order to detect a stroke in table tennis and assess its duration. This implementation is an extension of the baseline code provided by the Sport Task organizers~\cite{mediaeval/Martin21/baseline}.

\section{Approach}
\label{sec:approach}
Initially, we sought to use ball detection and tracking to perform stroke detection. The first implementation used the pretrained model TTNet~\cite{voeikov2020ttnet}. However, the model failed to adapt to the acquisition conditions from TTStroke-21~\cite{PeCBMI:2018}, on which the task is built upon, and no fine-tuning was possible since no ball coordinates are available in the provided annotations. Therefore we decided to train a model from scratch.
\par
In this section, we first present the preparation of the videos and then the model presenting the processed data. Both processes are depicted in Fig.~\ref{fig:model_fig}. Post processing is performed to form a final decision.

\subsection{Data Preparation}

In video content analysis, the motion of objects of interest between frames can be of significant interest in order to understand their evolution in space. As such, we decided to use optical flow as a modality to perform stroke detection. Inspired by~\cite{PeICIP19}, we decided to use DeepFlow method~\cite{deepflow} to compute the optical flow from consecutive frames. The optical flow is computed from frames resized to $320\times128$. This size was initially chosen to keep the ball at least two pixels big, as it has previously been done in~\cite{voeikov2020ttnet}. Both the RGB and optical flow frames are consecutively stacked in a tensor of length 75. As in~\cite{mediaeval/Martin21/baseline}, stroke detection is tackled as a classification problem with two classes: ``Stroke'' and ``Non-stroke''.

\subsection{Model}

As shown in figure~\ref{fig:model_fig}, our Two-Stream model is composed of two branches of the same length. Each branch is a succession of four blocks and each block is composed of a convolutional layer with $3\times3\times3$ filters, followed by a ReLU activation function, and a $2\times2\times2$ pooling layer. The output of each branch is then flattened and fed to a fully connected layer that outputs a feature vector of length $500$. Both feature vectors are then concatenated and fed into a final fully connected layer of length two to predict the ``Stroke'' and ``Non-storke'' classes. One branch takes RGB frames of the video and the other computed optical flow. The model is trained using a stochastic gradient descent method over $250$ epochs with a learning rate of $0.001$, a batch size of $10$, a weight decay of $0.005$, and a Nesterov momentum~\cite{sutskever2013importance} of $0.5$. The negative samples creation and input processing is the same as the baseline~\cite{mediaeval/Martin21/baseline}.

\subsection{Post Processing}

Our model classifies 75 consecutive frames. In order to create stroke segments over the whole video, we classify every 75 frames of the videos, which leads to applying a sliding window without overlap. If two consecutive segments are classified as stroke, the segments are fused to create only one stroke.

\section{Results and Analysis}
\label{sec:res}
The metrics for evaluating the detection performance are described in~\cite{mediaeval/Martin21/task}. Our approach reached a mean Average Precision (mAP) of $0.00124$ and a Global Intersection over Union (G-IoU) of $0.0700$. It falls behind the baseline which reaches respectively $0.0173$ and $0.144$. Our other attempts using early concatenation of the RGB and Optical Flow modalities - meaning an input of size $5\times320\times128$ in one branch model - or training method without shuffling of the data, reached even lesser performance.

\par

\begin{figure}
 \centering
 \includegraphics[width=.95\linewidth]{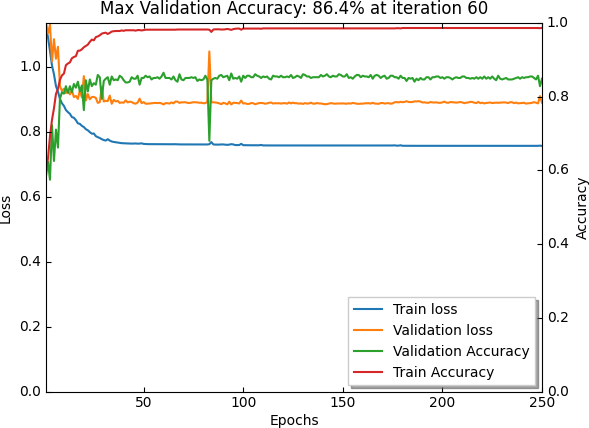}
 \caption{Training Process}
 \label{fig:train}
\end{figure}

Nevertheless, from a classification point of view, and according to the Fig.~\ref{fig:train}, our model learned the stroke features and can perform reasonable results when stroke boundaries are known: $86.4\%$ of accuracy on the validation set after only 60 epochs. Which may indicates that the main failure is coming from the post processing method.

Indeed, by looking at the stroke distribution across the different sets, see table~\ref{tab:hist}, we may notice how little the inferred stroke ratio is on the test set: 0.57 strokes for 1000 frames, whereas the stroke rate is 1.85 and 2.28 for 1000 frames in the training and validation sets. Furthermore, our post processing was not limited in term of stroke duration, leading to everlasting strokes: 4500 frames - meaning the fusions of 60 consecutive video segments. These
points indicate that our post-processing method can be improved.

\begin{table}
    \caption{Stroke concentration and duration in frame per set.}
    \begin{tabular}{|c|cccc|}
    \hline
    Set  & \# Strokes/1K frames & Mean  & Min  & Max \\
    \hline
    Train & 1.85    & $143.2\pm36.16$   & 52 & 296   \\
    Valid & 2.28    & $134.3\pm26.13$   & 72 & 292   \\
    Test  & 0.57    & $361.0\pm770.7$   & 75 & 4500  \\ 
    \hline
    \end{tabular}
    \label{tab:hist}
\end{table}


\par 

A better separation of the stroke may be reached by defining the event using ball tracking and the ball motion~\cite{calandre2021table}. This was our initial attempt, inspired by~\cite{voeikov2020ttnet}, but the available pretrained model considers a different point of view and was unable to adapt to the \texttt{TTStroke-21} videos point of view.


\section{Conclusion}
\label{sec:conclusion}

The Sports Video Task, and more specifically the stroke detection subtask, has proven to be challenging. Even if our implementation has learned to classify strokes, we were not able to outperform the baseline performance. We have underlined the importance of the post processing step through a stroke concentration and duration analysis. Furthermore, our failure to adapt a pretrained model on similar dataset, but with a different acquisition point of view, stresses the difficulty of the deep trained models to adapt to a change of scene, which is inherent to the fine-grained aspect of the classification subtask. As first time participants, we thought to tackle only one task to ease our submission. However, we now believe that a method tackling both the detection and classification may be the best for solving the Sport Video subtasks.

\bibliographystyle{ACM-Reference-Format}
\def\bibfont{\small} 
\bibliography{sigproc} 


\begin{thebibliography}{00}


\ifx \showCODEN    \undefined \def \showCODEN     #1{\unskip}     \fi
\ifx \showDOI      \undefined \def \showDOI       #1{#1}\fi
\ifx \showISBNx    \undefined \def \showISBNx     #1{\unskip}     \fi
\ifx \showISBNxiii \undefined \def \showISBNxiii  #1{\unskip}     \fi
\ifx \showISSN     \undefined \def \showISSN      #1{\unskip}     \fi
\ifx \showLCCN     \undefined \def \showLCCN      #1{\unskip}     \fi
\ifx \shownote     \undefined \def \shownote      #1{#1}          \fi
\ifx \showarticletitle \undefined \def \showarticletitle #1{#1}   \fi
\ifx \showURL      \undefined \def \showURL       {\relax}        \fi
\providecommand\bibfield[2]{#2}
\providecommand\bibinfo[2]{#2}
\providecommand\natexlab[1]{#1}
\providecommand\showeprint[2][]{arXiv:#2}

\bibitem[\protect\citeauthoryear{Ballan, Bertini, Del~Bimbo, Seidenari, and
  Serra}{Ballan et~al\mbox{.}}{2011}]%
        {ballan2011event}
\bibfield{author}{\bibinfo{person}{Lamberto Ballan}, \bibinfo{person}{Marco
  Bertini}, \bibinfo{person}{Alberto Del~Bimbo}, \bibinfo{person}{Lorenzo
  Seidenari}, {and} \bibinfo{person}{Giuseppe Serra}.}
  \bibinfo{year}{2011}\natexlab{}.
\newblock \showarticletitle{Event detection and recognition for semantic
  annotation of video}.
\newblock \bibinfo{journal}{{\em Multimedia tools and applications\/}}
  \bibinfo{volume}{51}, \bibinfo{number}{1} (\bibinfo{year}{2011}),
  \bibinfo{pages}{279--302}.
\newblock


\bibitem[\protect\citeauthoryear{Calandre, P{\'e}teri, Mascarilla, and
  Tremblais}{Calandre et~al\mbox{.}}{2021}]%
        {calandre2021table}
\bibfield{author}{\bibinfo{person}{Jordan Calandre}, \bibinfo{person}{Renaud
  P{\'e}teri}, \bibinfo{person}{Laurent Mascarilla}, {and}
  \bibinfo{person}{Benoit Tremblais}.} \bibinfo{year}{2021}\natexlab{}.
\newblock \showarticletitle{Table Tennis ball kinematic parameters estimation
  from non-intrusive single-view videos}. In \bibinfo{booktitle}{{\em 2021
  International Conference on Content-Based Multimedia Indexing (CBMI)}}. IEEE,
  \bibinfo{pages}{1--6}.
\newblock


\bibitem[\protect\citeauthoryear{Carreira and Zisserman}{Carreira and
  Zisserman}{2017}]%
        {carreira2017quo}
\bibfield{author}{\bibinfo{person}{Joao Carreira} {and} \bibinfo{person}{Andrew
  Zisserman}.} \bibinfo{year}{2017}\natexlab{}.
\newblock \showarticletitle{Quo vadis, action recognition? a new model and the
  kinetics dataset}. In \bibinfo{booktitle}{{\em proceedings of the IEEE
  Conference on Computer Vision and Pattern Recognition}}.
  \bibinfo{pages}{6299--6308}.
\newblock


\bibitem[\protect\citeauthoryear{Dhamsania and Ratanpara}{Dhamsania and
  Ratanpara}{2016}]%
        {dhamsania2016survey}
\bibfield{author}{\bibinfo{person}{Chandni~J Dhamsania} {and}
  \bibinfo{person}{Tushar~V Ratanpara}.} \bibinfo{year}{2016}\natexlab{}.
\newblock \showarticletitle{A survey on human action recognition from videos}.
  In \bibinfo{booktitle}{{\em 2016 online international conference on green
  engineering and technologies (IC-GET)}}. IEEE, \bibinfo{pages}{1--5}.
\newblock


\bibitem[\protect\citeauthoryear{Girshick}{Girshick}{2015}]%
        {girshick2015fast}
\bibfield{author}{\bibinfo{person}{Ross Girshick}.}
  \bibinfo{year}{2015}\natexlab{}.
\newblock \showarticletitle{Fast r-cnn}. In \bibinfo{booktitle}{{\em
  Proceedings of the IEEE international conference on computer vision}}.
  \bibinfo{pages}{1440--1448}.
\newblock


\bibitem[\protect\citeauthoryear{Hughes and Bartlett}{Hughes and
  Bartlett}{2002}]%
        {hughes2002use}
\bibfield{author}{\bibinfo{person}{Mike~D Hughes} {and}
  \bibinfo{person}{Roger~M Bartlett}.} \bibinfo{year}{2002}\natexlab{}.
\newblock \showarticletitle{The use of performance indicators in performance
  analysis}.
\newblock \bibinfo{journal}{{\em Journal of sports sciences\/}}
  \bibinfo{volume}{20}, \bibinfo{number}{10} (\bibinfo{year}{2002}),
  \bibinfo{pages}{739--754}.
\newblock


\bibitem[\protect\citeauthoryear{Khan and Pawar}{Khan and Pawar}{2015}]%
        {khan2015video}
\bibfield{author}{\bibinfo{person}{Yasmin~S Khan} {and}
  \bibinfo{person}{Soudamini Pawar}.} \bibinfo{year}{2015}\natexlab{}.
\newblock \showarticletitle{Video summarization: survey on event detection and
  summarization in soccer videos}.
\newblock \bibinfo{journal}{{\em International Journal of Advanced Computer
  Science and Applications\/}} \bibinfo{volume}{6}, \bibinfo{number}{11}
  (\bibinfo{year}{2015}), \bibinfo{pages}{256--259}.
\newblock


\bibitem[\protect\citeauthoryear{Koch, Zemel, Salakhutdinov,
  et~al\mbox{.}}{Koch et~al\mbox{.}}{2015}]%
        {koch2015siamese}
\bibfield{author}{\bibinfo{person}{Gregory Koch}, \bibinfo{person}{Richard
  Zemel}, \bibinfo{person}{Ruslan Salakhutdinov}, {and}
  \bibinfo{person}{others}.} \bibinfo{year}{2015}\natexlab{}.
\newblock \showarticletitle{Siamese neural networks for one-shot image
  recognition}. In \bibinfo{booktitle}{{\em ICML deep learning workshop}},
  Vol.~\bibinfo{volume}{2}. Lille.
\newblock


\bibitem[\protect\citeauthoryear{Krizhevsky, Sutskever, and Hinton}{Krizhevsky
  et~al\mbox{.}}{2012}]%
        {krizhevsky2012imagenet}
\bibfield{author}{\bibinfo{person}{Alex Krizhevsky}, \bibinfo{person}{Ilya
  Sutskever}, {and} \bibinfo{person}{Geoffrey~E Hinton}.}
  \bibinfo{year}{2012}\natexlab{}.
\newblock \showarticletitle{Imagenet classification with deep convolutional
  neural networks}.
\newblock \bibinfo{journal}{{\em Advances in neural information processing
  systems\/}}  \bibinfo{volume}{25} (\bibinfo{year}{2012}),
  \bibinfo{pages}{1097--1105}.
\newblock


\bibitem[\protect\citeauthoryear{Lees}{Lees}{2003}]%
        {lees2003science}
\bibfield{author}{\bibinfo{person}{Adrian Lees}.}
  \bibinfo{year}{2003}\natexlab{}.
\newblock \showarticletitle{Science and the major racket sports: a review}.
\newblock \bibinfo{journal}{{\em Journal of sports sciences\/}}
  \bibinfo{volume}{21}, \bibinfo{number}{9} (\bibinfo{year}{2003}),
  \bibinfo{pages}{707--732}.
\newblock


\bibitem[\protect\citeauthoryear{Martin}{Martin}{2021}]%
        {mediaeval/Martin21/baseline}
\bibfield{author}{\bibinfo{person}{Pierre{-}Etienne Martin}.}
  \bibinfo{year}{2021}\natexlab{}.
\newblock \showarticletitle{Spatio-Temporal CNN baseline method for the Sports
  Video Task of MediaEval 2021 benchmark}. In \bibinfo{booktitle}{{\em
  MediaEval}} {\em (\bibinfo{series}{{CEUR} Workshop Proceedings})}.
  \bibinfo{publisher}{CEUR-WS.org}.
\newblock


\bibitem[\protect\citeauthoryear{Martin, Benois{-}Pineau, Mansencal,
  P{\'{e}}teri, Mascarilla, Calandre, and Morlier}{Martin
  et~al\mbox{.}}{2021}]%
        {mediaeval/Martin21/task}
\bibfield{author}{\bibinfo{person}{Pierre{-}Etienne Martin},
  \bibinfo{person}{Jenny Benois{-}Pineau}, \bibinfo{person}{Boris Mansencal},
  \bibinfo{person}{Renaud P{\'{e}}teri}, \bibinfo{person}{Laurent Mascarilla},
  \bibinfo{person}{Jordan Calandre}, {and} \bibinfo{person}{Julien Morlier}.}
  \bibinfo{year}{2021}\natexlab{}.
\newblock \showarticletitle{Sports Video: Fine-Grained Action Detection and
  Classification of Table Tennis Strokes from videos for MediaEval 2021}. In
  \bibinfo{booktitle}{{\em MediaEval}} {\em (\bibinfo{series}{{CEUR} Workshop
  Proceedings})}. \bibinfo{publisher}{CEUR-WS.org}.
\newblock


\bibitem[\protect\citeauthoryear{Martin, Benois{-}Pineau, P{\'{e}}teri, and
  Morlier}{Martin et~al\mbox{.}}{2018}]%
        {PeCBMI:2018}
\bibfield{author}{\bibinfo{person}{Pierre{-}Etienne Martin},
  \bibinfo{person}{Jenny Benois{-}Pineau}, \bibinfo{person}{Renaud
  P{\'{e}}teri}, {and} \bibinfo{person}{Julien Morlier}.}
  \bibinfo{year}{2018}\natexlab{}.
\newblock \showarticletitle{Sport Action Recognition with Siamese
  Spatio-Temporal CNNs: Application to Table Tennis}. In
  \bibinfo{booktitle}{{\em {CBMI}}}. \bibinfo{publisher}{{IEEE}},
  \bibinfo{pages}{1--6}.
\newblock


\bibitem[\protect\citeauthoryear{Martin, Benois{-}Pineau, P{\'{e}}teri, and
  Morlier}{Martin et~al\mbox{.}}{2019}]%
        {PeICIP19}
\bibfield{author}{\bibinfo{person}{Pierre{-}Etienne Martin},
  \bibinfo{person}{Jenny Benois{-}Pineau}, \bibinfo{person}{Renaud
  P{\'{e}}teri}, {and} \bibinfo{person}{Julien Morlier}.}
  \bibinfo{year}{2019}\natexlab{}.
\newblock \showarticletitle{Optimal Choice of Motion Estimation Methods for
  Fine-Grained Action Classification with 3D Convolutional Networks}. In
  \bibinfo{booktitle}{{\em 2019 {IEEE} International Conference on Image
  Processing, {ICIP} 2019, Taipei, Taiwan, September 22-25, 2019}}.
  \bibinfo{publisher}{{IEEE}}, \bibinfo{pages}{554--558}.
\newblock
\showDOI{%
\url{https://doi.org/10.1109/ICIP.2019.8803780}}


\bibitem[\protect\citeauthoryear{Martin, Benois{-}Pineau, P{\'{e}}teri, and
  Morlier}{Martin et~al\mbox{.}}{2020a}]%
        {PeICPR20}
\bibfield{author}{\bibinfo{person}{Pierre{-}Etienne Martin},
  \bibinfo{person}{Jenny Benois{-}Pineau}, \bibinfo{person}{Renaud
  P{\'{e}}teri}, {and} \bibinfo{person}{Julien Morlier}.}
  \bibinfo{year}{2020}\natexlab{a}.
\newblock \showarticletitle{3D attention mechanisms in Twin Spatio-Temporal
  Convolutional Neural Networks. Application to action classification in videos
  of table tennis games.}. In \bibinfo{booktitle}{{\em 25th International
  Conference on Pattern Recognition (ICPR2020) - MiCo Milano Congress Center,
  Italy, 10-15 January 2021}}.
\newblock


\bibitem[\protect\citeauthoryear{Martin, Benois-Pineau, P{\'e}teri, and
  Morlier}{Martin et~al\mbox{.}}{2020b}]%
        {martin2020fine}
\bibfield{author}{\bibinfo{person}{Pierre-Etienne Martin},
  \bibinfo{person}{Jenny Benois-Pineau}, \bibinfo{person}{Renaud P{\'e}teri},
  {and} \bibinfo{person}{Julien Morlier}.} \bibinfo{year}{2020}\natexlab{b}.
\newblock \showarticletitle{Fine grained sport action recognition with twin
  spatio-temporal convolutional neural networks}.
\newblock \bibinfo{journal}{{\em Multimedia Tools and Applications\/}}
  \bibinfo{volume}{79}, \bibinfo{number}{27} (\bibinfo{year}{2020}),
  \bibinfo{pages}{20429--20447}.
\newblock


\bibitem[\protect\citeauthoryear{Myint, Wong, Dooley, and Hopgood}{Myint
  et~al\mbox{.}}{2015}]%
        {myint2015tracking}
\bibfield{author}{\bibinfo{person}{Hnin Myint}, \bibinfo{person}{Patrick Wong},
  \bibinfo{person}{Laurence Dooley}, {and} \bibinfo{person}{Adrian Hopgood}.}
  \bibinfo{year}{2015}\natexlab{}.
\newblock \showarticletitle{Tracking a table tennis ball for umpiring
  purposes}. In \bibinfo{booktitle}{{\em 2015 14th IAPR International
  Conference on Machine Vision Applications (MVA)}}. IEEE,
  \bibinfo{pages}{170--173}.
\newblock


\bibitem[\protect\citeauthoryear{Myint, Wong, Dooley, and Hopgood}{Myint
  et~al\mbox{.}}{2016}]%
        {myint2016tracking}
\bibfield{author}{\bibinfo{person}{Hnin Myint}, \bibinfo{person}{Patrick Wong},
  \bibinfo{person}{Laurence Dooley}, {and} \bibinfo{person}{Adrian Hopgood}.}
  \bibinfo{year}{2016}\natexlab{}.
\newblock \showarticletitle{Tracking a table tennis ball for umpiring purposes
  using a multi-agent system}.
\newblock  (\bibinfo{year}{2016}).
\newblock


\bibitem[\protect\citeauthoryear{Sutskever, Martens, Dahl, and
  Hinton}{Sutskever et~al\mbox{.}}{2013}]%
        {sutskever2013importance}
\bibfield{author}{\bibinfo{person}{Ilya Sutskever}, \bibinfo{person}{James
  Martens}, \bibinfo{person}{George Dahl}, {and} \bibinfo{person}{Geoffrey
  Hinton}.} \bibinfo{year}{2013}\natexlab{}.
\newblock \showarticletitle{On the importance of initialization and momentum in
  deep learning}. In \bibinfo{booktitle}{{\em International conference on
  machine learning}}. PMLR, \bibinfo{pages}{1139--1147}.
\newblock


\bibitem[\protect\citeauthoryear{Tamaki and Saito}{Tamaki and Saito}{2013}]%
        {tamaki2013reconstruction}
\bibfield{author}{\bibinfo{person}{Sho Tamaki} {and} \bibinfo{person}{Hideo
  Saito}.} \bibinfo{year}{2013}\natexlab{}.
\newblock \showarticletitle{Reconstruction of 3d trajectories for performance
  analysis in table tennis}. In \bibinfo{booktitle}{{\em Proceedings of the
  IEEE Conference on Computer Vision and Pattern Recognition Workshops}}.
  \bibinfo{pages}{1019--1026}.
\newblock


\bibitem[\protect\citeauthoryear{Voeikov, Falaleev, and Baikulov}{Voeikov
  et~al\mbox{.}}{2020}]%
        {voeikov2020ttnet}
\bibfield{author}{\bibinfo{person}{Roman Voeikov}, \bibinfo{person}{Nikolay
  Falaleev}, {and} \bibinfo{person}{Ruslan Baikulov}.}
  \bibinfo{year}{2020}\natexlab{}.
\newblock \showarticletitle{TTNet: Real-time temporal and spatial video
  analysis of table tennis}. In \bibinfo{booktitle}{{\em Proceedings of the
  IEEE/CVF Conference on Computer Vision and Pattern Recognition Workshops}}.
  \bibinfo{pages}{884--885}.
\newblock


\bibitem[\protect\citeauthoryear{Wang and Schmid}{Wang and Schmid}{2013}]%
        {wang2013action}
\bibfield{author}{\bibinfo{person}{Heng Wang} {and} \bibinfo{person}{Cordelia
  Schmid}.} \bibinfo{year}{2013}\natexlab{}.
\newblock \showarticletitle{Action recognition with improved trajectories}. In
  \bibinfo{booktitle}{{\em Proceedings of the IEEE international conference on
  computer vision}}. \bibinfo{pages}{3551--3558}.
\newblock


\bibitem[\protect\citeauthoryear{Weinzaepfel, Revaud, Harchaoui, and
  Schmid}{Weinzaepfel et~al\mbox{.}}{2013}]%
        {deepflow}
\bibfield{author}{\bibinfo{person}{Philippe Weinzaepfel},
  \bibinfo{person}{Jerome Revaud}, \bibinfo{person}{Zaid Harchaoui}, {and}
  \bibinfo{person}{Cordelia Schmid}.} \bibinfo{year}{2013}\natexlab{}.
\newblock \showarticletitle{DeepFlow: Large Displacement Optical Flow with Deep
  Matching}. In \bibinfo{booktitle}{{\em 2013 IEEE International Conference on
  Computer Vision}}. \bibinfo{pages}{1385--1392}.
\newblock
\showDOI{%
\url{https://doi.org/10.1109/ICCV.2013.175}}


\end{thebibliography}
\end{document}